\documentclass[sigconf, screen, nonacm]{acmart}

\AtBeginDocument{%
  }

\setcopyright{acmlicensed}
\copyrightyear{2018}
\acmYear{2018}
\acmDOI{XXXXXXX.XXXXXXX}
\acmConference[Conference acronym 'XX]{Make sure to enter the correct
  conference title from your rights confirmation email}{June 03--05,
  2018}{Woodstock, NY}
\acmISBN{978-1-4503-XXXX-X/2018/06}

\begin{document}

\title[One Shared LoRA Weight for MRI Reconstruction across Acceleration Factors]
  {One Shared LoRA Weight for MRI Reconstruction across Acceleration Factors}

\author{Zhiwei Zhao, Weikang Gong, Zhongnian Li\textsuperscript{*}, Xinzheng Xu}
\affiliation{%
  \institution{School of Computer Science and Technology / School of Artificial Intelligence, China University of Mining and Technology}
  \city{Xuzhou}
  \country{China}
}
\email{\{ts25170035a31ld, wkgong, zhongnianli, xxzheng\}@cumt.edu.cn}

\renewcommand{\shortauthors}{Zhao et al.}

\begin{abstract}
Accelerated MRI reconstruction recovers images from undersampled k-space. However, different acceleration factors produce distinct artifact patterns. Existing methods often train separate models for each factor, leading to poor cross-factor generalization and high training and storage costs. We propose Shared LoRA, a parameter-efficient framework that freezes the pretrained SHFormer backbone and trains a single shared set of LoRA adapters together with a lightweight gating network. During training, undersampled inputs are generated by randomly sampling acceleration factors and their corresponding sampling masks, enabling the shared adapters to learn reconstruction knowledge across factors. Given the acceleration factor, GateNet generates layer-wise coefficients to dynamically modulate the residual strength of each adapter. Experiments show that Shared LoRA achieves the best or competitive PSNR and SSIM across acceleration factors, while its trainable parameters account for only about 5.3\% of the total model parameters. Its performance at lower acceleration factors remains largely unaffected as the jointly trained factor set expands, and it generalizes stably to unseen neighboring factors.
\end{abstract}



\maketitle

\section{Introduction}

\begin{figure*}[t]
  \centering
  \includegraphics[width=\textwidth]{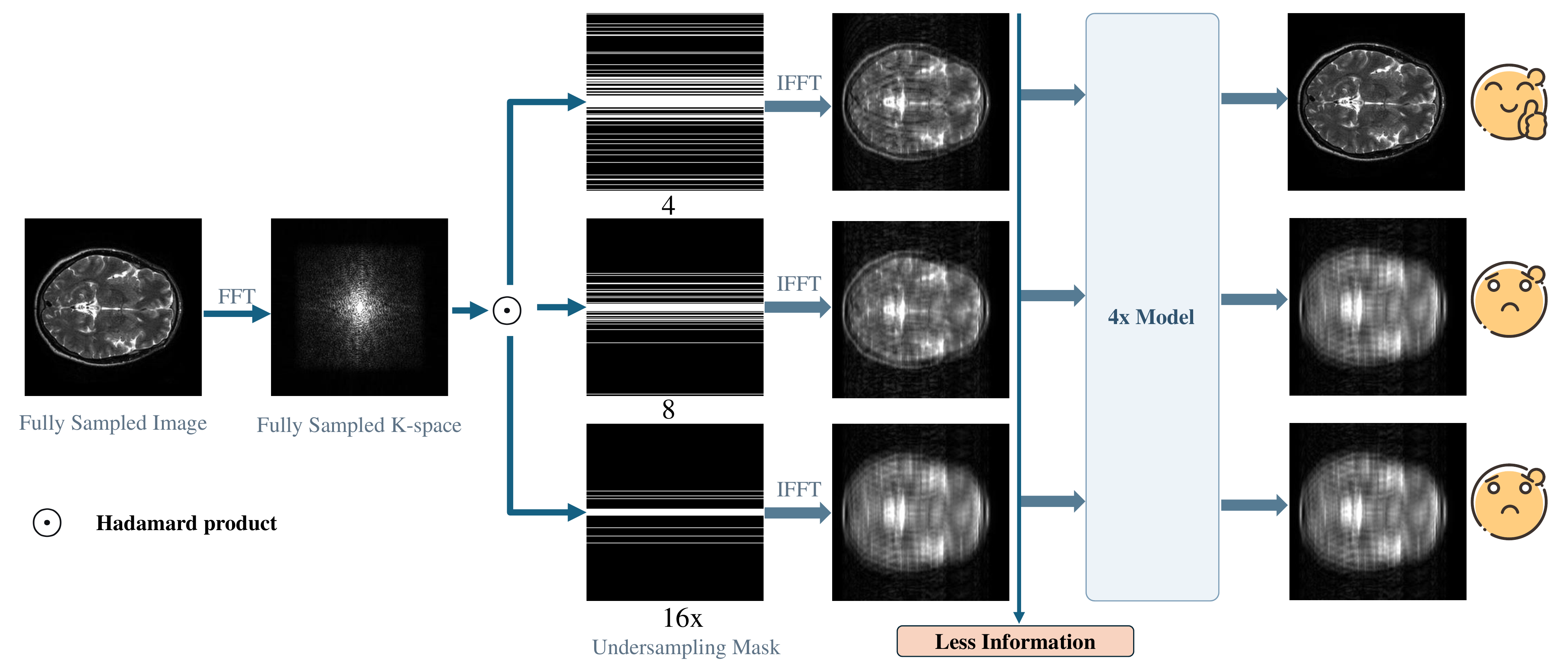}
  \caption{When the pre-trained model, optimized for a single acceleration factor, is directly applied to out-of-distribution undersampling ratios, its reconstruction quality collapses—exhibiting pronounced aliasing artifacts and structural blurring.}
  \label{fig:motivation}
\end{figure*}

Magnetic resonance imaging (MRI) provides high soft-tissue contrast and
multiparametric information, but measurements must be acquired sequentially
through repeated encoding steps in k-space, making the acquisition process
time-consuming. Accelerated MRI reconstruction addresses this problem by
undersampling k-space and using reconstruction algorithms to recover images
from incomplete measurements\cite{cukur2026mri,heckel2024deep}. However, most
deep reconstruction networks are optimized for only a single acceleration
factor and sampling pattern. Once the undersampling rate or sampling pattern
changes, k-space coverage and artifact characteristics change accordingly,
leading to residual artifacts, blurred details, or
overcorrection\cite{heckel2024deep,noordman2023complexities}, as shown in
Figure~\ref{fig:motivation}.

Although training an independent model for each factor achieves specialization,
the training, storage, and deployment costs increase multiplicatively.
Parameter-efficient fine-tuning (PEFT) reduces the number of trainable
parameters, yet factor-specific adapters still need to be stored and loaded
separately, and as more acceleration factors and sampling masks are supported, the need to maintain multiple adapters gradually offsets the storage and deployment advantages of parameter-efficient adaptation\cite{houlsby2019adapter,hu2022lora}.

To address these issues, we propose Shared LoRA, which adopts a pretrained
SHFormer\cite{ramanarayanan2025shformer} as the reconstruction backbone with
all parameters frozen, and inserts shared low-rank convolutional branches only
into selected convolutional layers of the encoder and decoder in SFCNN. The
residual outputs are modulated by gating according to the acceleration factor
and then added back to the backbone feature stream. GateNet is a lightweight gating network that takes the acceleration factor as input and generates layer-wise gate coefficients to modulate the residual output strength of each branch. In this way, a single set of adapter parameters covers multiple acceleration factors and generalizes to unseen neighboring factors. During training, a mixed
undersampled-input strategy randomly selects an acceleration factor and its
corresponding pre-generated undersampling mask to construct undersampled inputs
at different factors, enabling the same set of adapter parameters to reuse
shared reconstruction knowledge across different factors.

Extensive experiments show that Shared LoRA superior reconstruction quality across multiple acceleration factors with a single set
of shared adapter parameters, while alleviating task interference in joint
multi-factor training and generalizing stably to unseen neighboring factors.
Specifically, on IXI-T2, with a single set of adapter parameters covering
\(4\times\)--\(16\times\), the \(4\times\) PSNR reaches 42.17~dB, an improvement
of 0.61~dB over the strongest baseline, while the added parameters account for
only about 5.3\% of the total parameters. Our contributions are:

\begin{enumerate}
  \item We propose shared low-rank adaptation with an
    acceleration-factor-conditioned gating network (GateNet), which achieves
    reconstruction adaptation across multiple acceleration factors with a
    single set of adapter parameters and generalizes to unseen neighboring
    factors.
  \item We design a mixed undersampled-input training strategy across
    acceleration factors, enabling the same set of adapter parameters to learn
    shared reconstruction knowledge under different undersampling conditions
    and alleviating task interference in joint multi-factor training.
\end{enumerate}

\section{Related Work}

\subsection{Accelerated MRI Reconstruction}

\begin{figure*}[t]
  \centering
  \includegraphics[width=\textwidth]{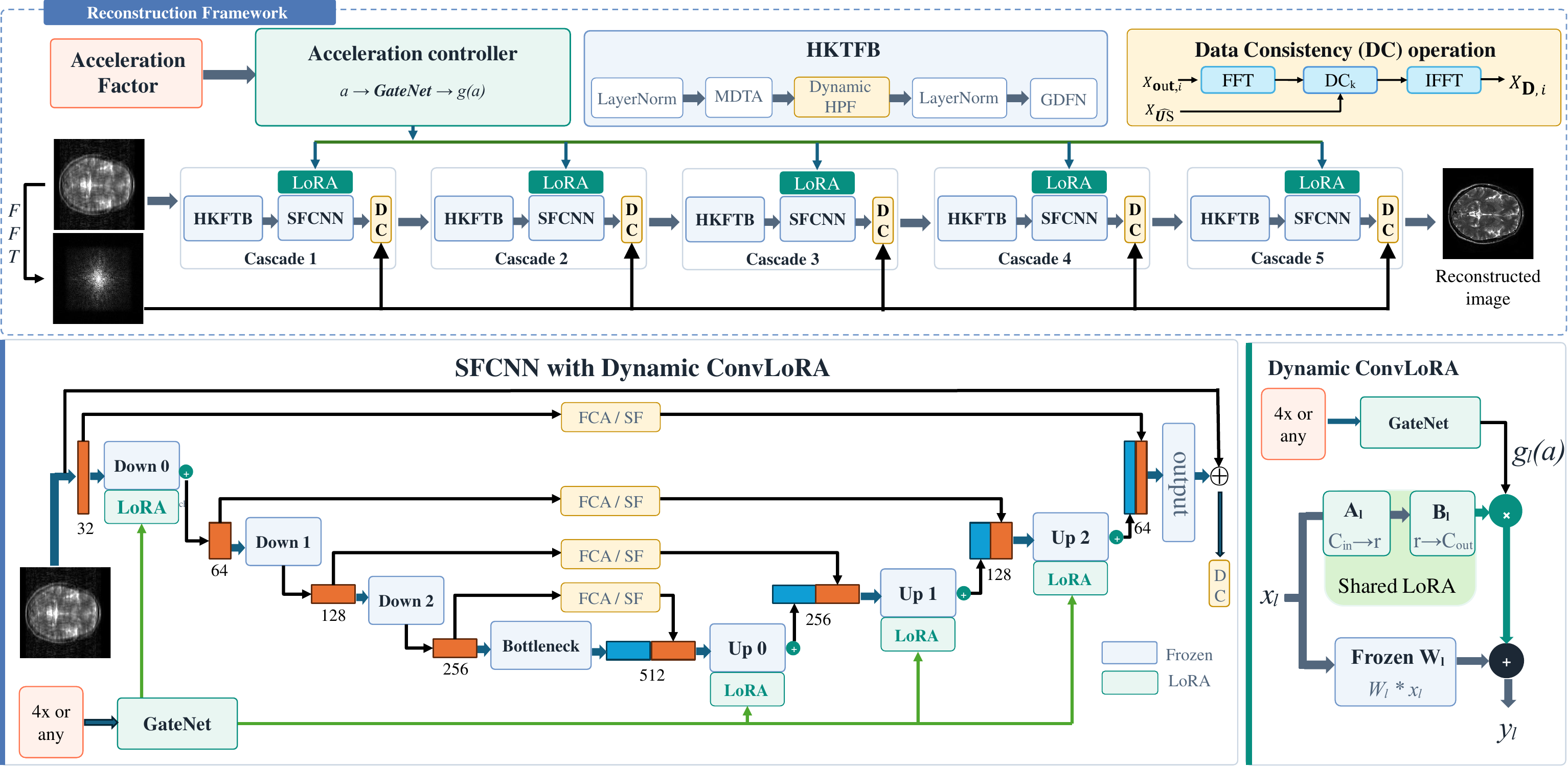}
  \caption{Architecture of the proposed reconstruction network. The upper panel shows the overall reconstruction framework, the lower-left panel details the internal structure of SFCNN, and the lower-right panel illustrates the operating mechanism of Dynamic ConvLoRA.}
  \label{fig:architecture}
\end{figure*}

Accelerated MRI reconstruction aims to recover images from undersampled
k-space. Early works model the imaging process as an explicit acquisition
operator and solve the inverse problem via iterative optimization with
handcrafted sparsity priors\cite{lustig2007sparse}. Deep learning methods
instead learn reconstruction priors from paired data\cite{eo2018kikinet,yang2018dagan}. Cascaded and unrolled
networks unfold optimization iterations into network layers and alternate
learned updates with data-consistency operations, preserving physical fidelity
while gaining expressiveness\cite{hammernik2018variational,schlemper2018deep,aggarwal2019modl,sriram2020varnet}.
Plug-and-play frameworks couple explicit regularizers with pretrained deep
denoisers, yielding model-driven reconstruction without additional
training\cite{wang2025pnp}. To avoid the need for fully sampled labels,
self-supervised approaches partition the acquired k-space into disjoint subsets
for network training and loss supervision, or embed self-supervised denoisers
directly into unrolled networks\cite{yaman2020self,huang2024dured}. Moreover,
systematic robustness analyses show that shifts in sampling pattern, anatomy,
or acceleration rate markedly degrade reconstruction
quality\cite{heckel2024deep,noordman2023complexities}. However, most of these
methods are optimized for a single acceleration factor. Expanding the factor
set typically requires training and maintaining one full model per factor,
incurring multiplicative costs and poor cross-factor generalization.

\subsection{Parameter-Efficient and Condition-Adaptive Reconstruction}

Beyond conventional architectures, attention-based and hybrid
convolution--attention models capture long-range dependencies and multiscale
structure\cite{qiu2024multicontrast,guo2024reconformer}, while frequency-domain methods decompose
images into high- and low-frequency components via Fourier
operations\cite{yi2023fmt}. SHFormer, the backbone adopted in this work,
couples dynamic spectral-filtering convolution with a high-pass-kernel-
generation Transformer, enhancing high-frequency detail recovery and
generalization to heterogeneous, unseen imaging
domains\cite{ramanarayanan2025shformer}.

Parameter-efficient adaptation (PEFT) freezes such a pretrained backbone and
learns only a small set of task-related parameters, typically in the form of
bottleneck adapters, low-rank weight updates, and their convolutional
variants\cite{houlsby2019adapter,hu2022lora,yeh2024navigating,chen2022adaptformer}. Building on
this, condition-aware methods further encode acquisition conditions into the
model. Hierarchical feature adapters adapt a shared backbone across
centers\cite{xu2026hieradaptmr}, prompt-guided models steer a universal network
toward different sampling trajectories\cite{lyu2024upcmr}, and federated
learning with test-time personalization adapts a global model to a target
center\cite{geng2026ttpssfl}. However, most parameter-efficient methods still
bind adapters to specific factors: each new factor demands an additional
adapter, so the more factors are supported, the more the repeated adapters
erode the storage and operational advantages.

\section{Method}

\subsection{Method Overview}

Shared LoRA builds on the pretrained SHFormer backbone and follows the
parameter-efficient paradigm of freezing the backbone while training only
lightweight adapters. All SHFormer parameters are frozen, and only the shared
low-rank branches and the gating network GateNet are optimized. The shared
low-rank branches are inserted into selected convolutional layers of the
encoder and decoder to provide low-rank residual updates. GateNet, conditioned
on the acceleration factor, generates layer-wise gate coefficients that
dynamically modulate the residual strength of each branch, allowing a single
set of adapter parameters to adapt to different undersampling conditions.
During training, a fully sampled image is combined with a randomly selected
acceleration factor and its corresponding undersampling mask to construct a
zero-filled undersampled input. During inference, the model reconstructs the
image from the zero-filled image generated from the observed undersampled data
and feeds the target acceleration factor to GateNet to produce the
corresponding gates.The architecture of Shared LoRA is shown in Figure~\ref{fig:architecture}.

\subsection{Shared LoRA in SFCNN}

We insert shared trainable low-rank convolutional branches into selected
convolutional layers of the frozen SHFormer backbone. Shared LoRA is applied to
selected convolutional layers in the SFCNN encoder and decoder paths. The
original reconstruction path and pretrained data-consistency operations remain
frozen, while the low-rank branches provide additional residual updates at the
selected layers. For the \(l\)-th adapted convolutional layer, the low-rank
residual is computed as
\begin{equation}
  \Delta y_l=\frac{\alpha_l}{r_l}B_l\left(A_l(x_l)\right).
  \label{eq:convlora-residual}
\end{equation}
Here, \(x_l\) is the input feature, \(A_l\) and \(B_l\) are the
dimension-reduction and dimension-expansion convolutional operators,
respectively, \(r_l\) is the low-rank dimension, \(\alpha_l\) is the scaling
coefficient, and \(\Delta y_l\) is the LoRA residual.

During training, the pretrained SHFormer parameters \(\theta\) are frozen. Only
the LoRA parameters \(\phi\) and GateNet parameters \(\psi\) are optimized:
\begin{equation}
  \Theta_{\mathrm{train}}=\{\phi,\psi\}.
  \label{eq:trainable-parameters}
\end{equation}
Here, \(\phi\) denotes all LoRA parameters in the target convolutional layers,
and \(\psi\) denotes the GateNet parameters.

\subsection{Acceleration-Factor-Conditioned Gating}
\begin{figure*}[t]
  \centering
  \includegraphics[width=\textwidth]{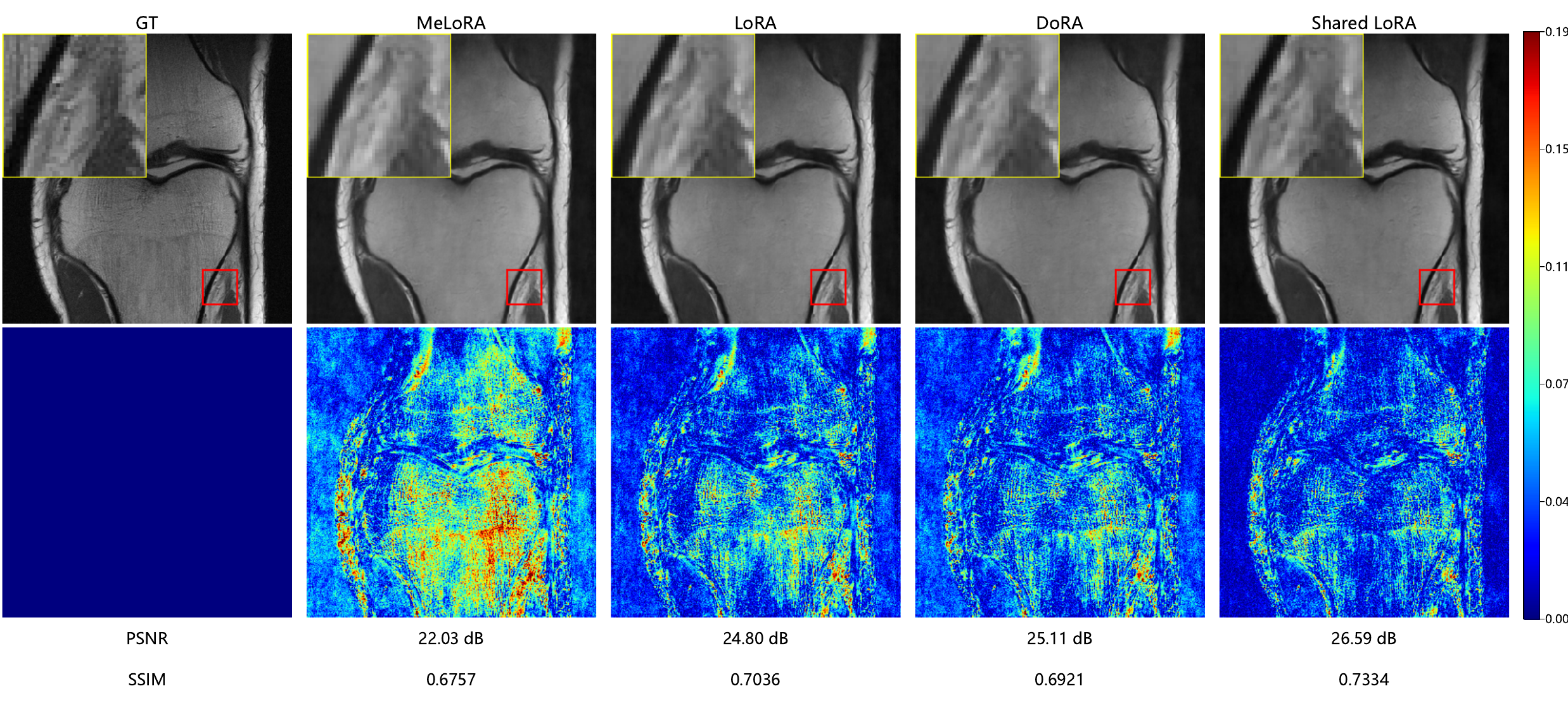}
  \caption{Qualitative comparison on FastMRI. Columns: GT, MeLoRA, LoRA, DoRA, Shared LoRA. Top row: center slices with ROI (red box, inset); bottom row: absolute-error maps (common color scale). PSNR/SSIM listed below.}
  \label{fig:reconstruction-residual}
\end{figure*}

Different acceleration factors produce different levels of information loss and
artifact severity, and changes in the undersampling pattern lead to artifact
mismatch that degrades the pretrained model. Moreover, in joint multi-factor
training, the shared LoRA branches tend to be dominated by the large gradients
of high-loss factors, biasing the model toward difficult samples and weakening
generalization at lower factors. Since a fixed residual strength cannot adapt
to different levels of information loss, we resort to factor-conditioned
modulation.

GateNet takes the scalar acceleration factor \(a\) as input and generates the
layer-wise gate vector
\begin{equation}
  \mathbf{g}(a)=G_{\psi}(a)\in\mathbb{R}^{D}.
  \label{eq:gate-vector}
\end{equation}
Here, \(\mathbf{g}(a)\) denotes the gate vector, \(D\) denotes its length, and
\(N\) denotes the number of adapted low-rank residual branches. For single
gating, \(D=N\). Within each batch, the same gate vector corresponding to the
batch acceleration factor is used for all samples.

For the \(l\)-th adapted layer, the gate \(g_l(a)\) scales only the residual
contribution of the low-rank branch:
\begin{equation}
  y_l=W_l*x_l+g_l(a)\Delta y_l.
  \label{eq:single-gate-output}
\end{equation}
The gate scales only the low-rank branch: the gate coefficient \(g_l(a)\)
modulates only the residual term \(g_l(a)\Delta y_l\), while the backbone
convolution term \(W_l*x_l\) is left untouched. GateNet is implemented as a
lightweight MLP with the structure \(1\rightarrow64\rightarrow64\rightarrow D\),
whose final layer is initialized to produce unit gates, so the low-rank
residuals are not rescaled at the start of training and the model starts
smoothly from the pretrained backbone.

\subsection{Factor-Randomized Input and Optimization}

During training, a fully sampled target image is combined with a randomly
selected acceleration factor and its corresponding undersampling mask to
generate the undersampled input:
\begin{equation}
  x_{\mathrm{us}}^{a}
  =
  \left|
  \mathcal{F}^{-1}
  \!\left(
  \mathcal{F}(x_{\mathrm{fs}})\odot M_a
  \right)
  \right|.
  \label{eq:mixed-input}
\end{equation}
Here, \(x_{\mathrm{fs}}\) is the fully sampled target, \(M_a\) is the mask
associated with factor \(a\), and \(x_{\mathrm{us}}^{a}\) is the generated
undersampled image.

During training, each mini-batch randomly selects one acceleration factor from
the training set. The selected factor is used to generate the undersampled input
and is simultaneously provided to GateNet, so all samples in the batch share the
same conditioning factor. The factor changes across mini-batches and training
iterations.

The image-domain L1 loss is defined as
\begin{equation}
  \mathcal{L}_{\mathrm{img}}
  =
  \frac{1}{BHW}
  \left\|\hat{x}-x\right\|_1.
  \label{eq:image-l1}
\end{equation}
Here, \(x\) and \(\hat{x}\) denote the fully sampled target and reconstructed
image batches, respectively. The backbone remains frozen, while \(\phi\) and
\(\psi\) are updated by the image-domain L1 loss. During inference, the target
acceleration factor is provided to GateNet, which generates the corresponding
gate vector for the shared adapter.

\subsection{Parameter Efficiency}
\begin{table*}[t]
  \caption{Comparison of Shared LoRA with other LoRA-based methods on the IXI dataset under input acceleration configurations including 16$\times$ and 32$\times$}
  \label{tab:ixi-comparison-other}
  \centering
  \scriptsize
  \renewcommand{\arraystretch}{1.15}
  \begin{tabular*}{\textwidth}{@{\extracolsep{\fill}}llcccccc@{}}
    \toprule
    Method & Input acceleration
      & \multicolumn{2}{c}{4$\times$}
      & \multicolumn{2}{c}{8$\times$}
      & \multicolumn{2}{c}{16$\times$} \\
    & & PSNR & SSIM & PSNR & SSIM & PSNR & SSIM \\
    \midrule
    LoRA & 4$\times$, 8$\times$, 16$\times$
      & $41.56 \pm 5.427$ & $0.9863 \pm 0.008895$
      & $33.88 \pm 5.390$ & $0.9519 \pm 0.02158$
      & $29.76 \pm 5.418$ & $0.8937 \pm 0.04039$ \\
    MeLoRA & 4$\times$, 8$\times$, 16$\times$
      & $41.47 \pm 5.379$ & $0.9860 \pm 0.008954$
      & $33.80 \pm 5.416$ & $0.9511 \pm 0.02194$
      & $29.66 \pm 5.442$ & $0.8917 \pm 0.04121$ \\
    DoRA & 4$\times$, 8$\times$, 16$\times$
      & $41.55 \pm 5.484$ & $0.9864 \pm 0.009053$
      & $33.95 \pm 5.365$ & $0.9526 \pm 0.02119$
      & $29.87 \pm 5.391$ & $0.8958 \pm 0.03956$ \\
    \textbf{Shared LoRA} & 4$\times$, 8$\times$, 16$\times$
      & \textbf{$42.17 \pm 5.662$} & \textbf{$0.9881 \pm 0.008563$}
      & \textbf{$34.01 \pm 5.420$} & \textbf{$0.9532 \pm 0.02138$}
      & \textbf{$29.94 \pm 5.406$} & \textbf{$0.8970 \pm 0.03952$} \\
    \midrule
    LoRA & 4$\times$, 8$\times$, 16$\times$, 32$\times$
      & $41.20 \pm 5.247$ & $0.9851 \pm 0.009176$
      & $33.56 \pm 5.430$ & $0.9481 \pm 0.02317$
      & $29.48 \pm 5.431$ & $0.8857 \pm 0.04339$ \\
    MeLoRA & 4$\times$, 8$\times$, 16$\times$, 32$\times$
      & $41.13 \pm 5.213$ & $0.9847 \pm 0.009178$
      & $33.49 \pm 5.437$ & $0.9474 \pm 0.02318$
      & $29.41 \pm 5.433$ & $0.8843 \pm 0.04347$ \\
    DoRA & 4$\times$, 8$\times$, 16$\times$, 32$\times$
      & $41.14 \pm 5.258$ & $0.9849 \pm 0.009231$
      & $33.6 \pm 5.418$ & $0.9485 \pm 0.02299$
      & $29.56 \pm 5.411$ & $0.8871 \pm 0.04254$ \\
    \textbf{Shared LoRA} & 4$\times$, 8$\times$, 16$\times$, 32$\times$
      & \textbf{$42.13 \pm 5.672$} & \textbf{$0.9880 \pm 0.008600$}
      & \textbf{$33.67 \pm 5.434$} & \textbf{$0.9497 \pm 0.02257$}
      & \textbf{$29.57 \pm 5.424$} & \textbf{$0.8874 \pm 0.04245$} \\
    \bottomrule
  \end{tabular*}
\end{table*}

\begin{table*}[t]
  \caption{Comparison of Shared LoRA with other LoRA-based methods on the fastMRI dataset under different input acceleration configurations}
  \label{tab:fastmri-comparison}
  \centering
  \scriptsize
  \renewcommand{\arraystretch}{1.15}
  \begin{tabular*}{\textwidth}{@{\extracolsep{\fill}}llcccccc@{}}
    \toprule
    Method & Input acceleration
      & \multicolumn{2}{c}{8$\times$}
      & \multicolumn{2}{c}{16$\times$}
      & \multicolumn{2}{c}{32$\times$} \\
    & & PSNR & SSIM & PSNR & SSIM & PSNR & SSIM \\
    \midrule
    LoRA & 8$\times$, 16$\times$, 32$\times$
      & $36.56 \pm 3.215$ & $0.9307 \pm 0.06551$
      & $31.21 \pm 3.809$ & $0.8593 \pm 0.07612$
      & $27.87 \pm 4.607$ & $0.7409 \pm 0.07749$ \\
    MeLoRA & 8$\times$, 16$\times$, 32$\times$
      & $36.45 \pm 3.153$ & $0.9295 \pm 0.06505$
      & $31.07 \pm 3.886$ & $0.8567 \pm 0.07505$
      & $27.72 \pm 4.778$ & $0.7350 \pm 0.07688$ \\
    DoRA & 8$\times$, 16$\times$, 32$\times$
      & $36.58 \pm 3.222$ & $0.9306 \pm 0.06555$
      & $31.27 \pm 3.847$ & $0.8603 \pm 0.07653$
      & $27.99 \pm 4.641$ & $0.7445 \pm 0.07791$ \\
    \textbf{Shared LoRA} & 8$\times$, 16$\times$, 32$\times$
      & \textbf{$37.23 \pm 3.570$} & \textbf{$0.9368 \pm 0.06604$}
      & \textbf{$31.42 \pm 3.915$} & \textbf{$0.8628 \pm 0.07705$}
      & \textbf{$28.24 \pm 4.466$} & \textbf{$0.7543 \pm 0.07943$} \\
    \bottomrule
  \end{tabular*}
\end{table*}

\begin{table}[t]
  \caption{Cross-acceleration generalization of Shared LoRA and LoRA under different training and test factors}
  \label{tab:combined-generalization}
  \centering
  \scriptsize
  \renewcommand{\arraystretch}{1.12}
  \begin{tabular*}{\linewidth}{@{\extracolsep{\fill}}ll lcc@{}}
    \toprule
    Method & \multicolumn{2}{c}{Input Acceleration factor} & PSNR & SSIM \\
    \midrule

    Shared LoRA
      & $4\times$, $8\times$, $16\times$ & $3\times$
      & $42.62 \pm 4.810$ & $0.9836 \pm 0.008812$ \\

    Shared LoRA
      & $4\times$, $8\times$, $16\times$ & $5\times$
      & $38.95 \pm 5.471$ & $0.9732 \pm 0.01523$ \\

    Shared LoRA
      & $4\times$, $8\times$, $16\times$ & $10\times$
      & $31.46 \pm 5.400$ & $0.9160 \pm 0.03304$ \\

    \midrule

    LoRA
      & $4\times$, $8\times$ & $10\times$
      & $31.05 \pm 5.345$ & $0.9094 \pm 0.03588$ \\

    LoRA
      & $4\times$, $8\times$ & $32\times$
      & $24.47 \pm 5.404$ & $0.6851 \pm 0.09737$ \\

    Shared LoRA
      & $4\times$, $8\times$ & $10\times$
      & $30.62 \pm 5.341$ & $0.9011 \pm 0.03748$ \\

    Shared LoRA
      & $4\times$, $8\times$ & $32\times$
      & $23.67 \pm 4.117$ & $0.5675 \pm 0.1027$ \\

    \bottomrule
  \end{tabular*}
\end{table}

The additional parameter count of one low-rank convolutional branch is
\begin{equation}
  P_{\mathrm{LoRA},l}=rC_{\mathrm{in}}k^2+C_{\mathrm{out}}r.
  \label{eq:lora-parameter-count}
\end{equation}
Here, \(C_{\mathrm{in}}\) is the number of input channels, \(C_{\mathrm{out}}\)
is the number of output channels, \(k\) is the kernel size, and \(r\) is the
rank. The number of convolutional parameters injected into SHFormer is
\begin{equation}
  P_{\mathrm{full},l}=C_{\mathrm{out}}C_{\mathrm{in}}k^2.
  \label{eq:full-convolution-parameter-count}
\end{equation}
The additional parameters introduced by GateNet are
\begin{equation}
  P_{\mathrm{Gate}}=(1+1)h+(h+1)h+(h+1)D_g.
  \label{eq:gatenet-parameter-count}
\end{equation}
Here, \(h\) is the hidden width and \(D_g\) is the output dimension. The term
\((1+1)h\) corresponds to the parameters from the input layer to the hidden
layer, \((h+1)h\) corresponds to the parameters between hidden layers, and
\((h+1)D_g\) corresponds to the parameters from the hidden layer to the output
layer.

To illustrate the cost of the added parameters relative to the original
network, we compare a single low-rank branch with the full convolution of its
injected layer; the parameter ratio is
\begin{equation}
  \frac{P_{\mathrm{LoRA},l}}{P_{\mathrm{full},l}}
  =
  \frac{r\,C_{\mathrm{in}}k^2 + C_{\mathrm{out}} r}
       {C_{\mathrm{out}}C_{\mathrm{in}}k^2}.
  \label{eq:parameter-ratio}
\end{equation}
Under the \(r=8\), \(k=3\) configuration, the parameter ratio of the low-rank
branch to the full convolution decreases inversely with the channel count \(C\),
so the deeper and wider the layer, the lower the cost.

Because the backbone is frozen and the low-rank branches are shared across
acceleration factors, the method requires only one shared adapter instead of
one adapter per factor. With 40 injected layers in total and GateNet, the
trainable adapter parameters account for approximately \(5.3\%\) of the total
model parameters, and the FLOPs increase is bounded by the order of \(r/C\).
Since the GateNet input is a batch-shared scalar, its computational overhead is
negligible.

\section{Experimental}

\subsection{Datasets}

We conduct experiments on T2-weighted brain MRI from IXI and knee MRI from
fastMRI. IXI-T2 is downloaded from the official IXI website%
\footnote{\url{https://brain-development.org/ixi-dataset/}} and contains 580
three-dimensional volumes. The data are min--max normalized to \([0,1]\) and
split at a ratio of \(8:1:1\) into 464 training, 58 validation, and 58 test
volumes.

For fastMRI Knee, we use the first 500 volumes from the\linebreak 
\texttt{singlecoil\_train} subset of \path{btoto3/fastmri-dl} on Hugging Face%
\footnote{\url{https://huggingface.co/datasets/btoto3/fastmri-dl/tree/main/singlecoil_train}}.
Fully sampled targets are read from \texttt{reconstruction\_rss}. The data
are min--max normalized to \([0,1]\) and split at a ratio of \(8:1:1\) into
400 training, 50 validation, and 50 test volumes. The training and validation set contains
only fully sampled target images, while the test sets
additionally contain undersampled images and undersampled k-space.

\subsection{Implementation Details}
\begin{table*}[t]
  \caption{Ablation study of LoRA insertion settings on the IXI dataset under 4$\times$, 8$\times$, and 16$\times$ acceleration factors}
  \label{tab:ixi-lora-position-ablation}
  \centering
  \scriptsize
  \setlength{\fboxsep}{1.6pt}
  \providecommand{\checkedbox}{\ensuremath{\checkmark}}
  \providecommand{\uncheckedbox}{}
  \renewcommand{\arraystretch}{1.08}
  \begin{tabular*}{\textwidth}{@{\extracolsep{\fill}}ccc@{\hspace{1.1em}}ccccccc@{}}
    \toprule
    \multicolumn{3}{c}{LoRA insertion position}
      & \multicolumn{2}{c}{4$\times$}
      & \multicolumn{2}{c}{8$\times$}
      & \multicolumn{2}{c}{16$\times$}
      & Added params \\
    Decoder & \shortstack{First encoder\\layer} & Encoder
      & PSNR & SSIM & PSNR & SSIM & PSNR & SSIM & \#params \\
    \midrule
    \checkedbox & \uncheckedbox & \uncheckedbox
      & $42.13 \pm 5.660$ & $0.9880 \pm 0.008577$
      & $33.90 \pm 5.424$ & $0.9520 \pm 0.02177$
      & $29.80 \pm 5.420$ & $0.8942 \pm 0.04058$
      & $223838$ \\
    \checkedbox & \checkedbox & \uncheckedbox
      & $42.17 \pm 5.662$ & $0.9881 \pm 0.008563$
      & $34.01 \pm 5.420$ & $0.9532 \pm 0.02138$
      & $29.94 \pm 5.406$ & $0.8970 \pm 0.03952$
      & $238928$ \\
    \checkedbox & \checkedbox & \checkedbox
      & $42.22 \pm 5.663$ & $0.9882 \pm 0.008493$
      & $34.30 \pm 5.389$ & $0.9563 \pm 0.02015$
      & $30.38 \pm 5.373$ & $0.9066 \pm 0.03612$
      & $359268$ \\
    \bottomrule
  \end{tabular*}
\end{table*}

\begin{table*}[t]
  \caption{Ablation study of LoRA insertion settings on the fastMRI dataset under 8$\times$, 16$\times$, and 32$\times$ acceleration factors}
  \label{tab:fastmri-lora-position-ablation}
  \centering
  \scriptsize
  \setlength{\fboxsep}{1.6pt}
  \providecommand{\checkedbox}{\ensuremath{\checkmark}}
  \providecommand{\uncheckedbox}{}
  \renewcommand{\arraystretch}{1.08}
  \begin{tabular*}{\textwidth}{@{\extracolsep{\fill}}ccc@{\hspace{1.1em}}ccccccc@{}}
    \toprule
    \multicolumn{3}{c}{LoRA insertion position}
      & \multicolumn{2}{c}{8$\times$}
      & \multicolumn{2}{c}{16$\times$}
      & \multicolumn{2}{c}{32$\times$}
      & Added params \\
    Decoder & \shortstack{First encoder\\layer} & Encoder
      & PSNR & SSIM & PSNR & SSIM & PSNR & SSIM & \#params \\
    \midrule
    \checkedbox & \uncheckedbox & \uncheckedbox
      & $37.21 \pm 3.572$ & $0.9367 \pm 0.06620$
      & $31.17 \pm 4.004$ & $0.8580 \pm 0.07624$
      & $27.92 \pm 4.637$ & $0.7419 \pm 0.07715$
      & $223{,}838$ \\
    \checkedbox & \checkedbox & \uncheckedbox
      & $37.23 \pm 3.570$ & $0.9368 \pm 0.06604$
      & $31.42 \pm 3.915$ & $0.8628 \pm 0.07705$
      & $28.24 \pm 4.466$ & $0.7543 \pm 0.07943$
      & $238{,}928$ \\
    \checkedbox & \checkedbox & \checkedbox
      & $37.30 \pm 3.639$ & $0.9374 \pm 0.06651$
      & $31.81 \pm 3.672$ & $0.8712 \pm 0.07645$
      & $28.77 \pm 4.216$ & $0.7744 \pm 0.07898$
      & $359{,}268$ \\
    \bottomrule
  \end{tabular*}
\end{table*}

Cartesian masks are used for IXI-T2 and Gaussian masks for fastMRI Knee. The
pretrained SHFormer is initialized from models trained at \(4\times\) and
\(8\times\), respectively. We set the LoRA rank to 8, the scaling coefficient
to 16, and the dropout rate to 0.05. The batch sizes are 5 and 4, with 3 and 10
training epochs for IXI-T2 and fastMRI Knee, respectively. The backbone uses
Adam during pretraining, while the adapters use AdamW. Reconstruction quality
is evaluated using PSNR and SSIM.

\subsection{Experimental Results}

\subsubsection{Comparative Experiments}

We conduct comparative experiments on two datasets, IXI and fastMRI, to
evaluate multiple adaptation methods. All experiments use
the same reconstruction backbone and change only the adaptation method. The
proposed method is compared with several other adaptation methods, and the
results are summarized in Tables~\ref{tab:ixi-comparison-other} and~\ref{tab:fastmri-comparison}. Specifically, we use the
LoCon~\cite{yeh2024navigating} method to adapt three comparison methods---
LoRA~\cite{hu2022lora}, DoRA~\cite{pmlr-v235-liu24bn}, and
MELoRA~\cite{ren2024melora}---from linear layers to convolutional layers by
replacing their matrix-form low-rank updates with equivalent low-rank
convolutional-kernel adaptations.The visual comparison is shown in Figure~\ref{fig:reconstruction-residual}.

On the IXI-T2 dataset (taking the \(4\times\)/\(8\times\)/\(16\times\)
mixed-input configuration as an example), Shared LoRA achieves 42.17~dB PSNR at
\(4\times\), an improvement of 0.61~dB over the strongest baseline, with the
highest SSIM (0.9881) as well. At \(8\times\) and \(16\times\), it remains on
par with the strongest baseline (34.01 versus 34.05~dB and 29.94 versus
29.93~dB). On fastMRI Knee (taking the \(8\times\)/\(16\times\)/\(32\times\)
configuration as an example), Shared LoRA achieves 37.23~dB PSNR at \(8\times\),
an improvement of 0.60~dB over the strongest baseline, with the highest SSIM
(0.9368) as well. at \(16\times\) and \(32\times\), it stays on par with or
close to the strongest baseline (31.42 versus 31.51~dB and 28.24 versus
28.24~dB). These results indicate that dynamic gating improves adaptation to
different levels of undersampling, allowing Shared LoRA to accommodate
different acceleration factors and alleviate task interference during joint
multi-factor training.

\subsubsection{Generalization Performance}

To evaluate generalization to unseen acceleration factors, we train on IXI
using undersampled data at \(4\times\), \(8\times\), and \(16\times\), and test
at factors \(3\times\), \(5\times\), and \(10\times\). We additionally
compare Shared LoRA and LoRA trained on \(4\times\) and \(8\times\) data. The
results appear in Table~\ref{tab:combined-generalization}.

At inference on the unseen factors \(3\times\), \(5\times\), and \(10\times\),
Shared LoRA maintains reconstruction performance within a reasonable range,
achieving PSNR values of 42.62, 38.95, and 31.46~dB, respectively, without
obvious degradation relative to the seen factors. More specifically, the
\(5\times\) and \(10\times\) factors lie within the interval defined by the
seen factors \(4\times\), \(8\times\), and \(16\times\), and their results
follow the expected interpolation trend: performance at \(5\times\) falls
between the \(4\times\) and \(8\times\) results, while performance at
\(10\times\) falls between the \(8\times\) and \(16\times\) results. Although
\(3\times\) lies outside the seen interval, it also remains stable without an
observable performance drop. These results suggest that GateNet learns gate estimates from the discrete training factors and interpolates them smoothly to unseen neighboring factors, supplying appropriate residual-strength modulation and enabling Shared LoRA to generalize stably across the factor range.

\subsubsection{Ablation of LoRA Insertion Locations}

To investigate the effect of LoRA insertion locations on reconstruction
performance, we vary only the locations and numbers of LoRA modules in the
encoder and decoder of SHFormer, comparing three insertion configurations. On
IXI, the adopted configuration lags full-layer injection by only 0.05, 0.29,
and 0.44~dB in PSNR at \(4\times\), \(8\times\), and \(16\times\),
respectively, while improving over decoder-only injection by 0.04, 0.11, and
0.14~dB, so the three configurations achieve nearly identical reconstruction
quality. However, full-layer injection adds 359268 parameters, about 1.5
times the 238928 of the adopted configuration, which itself adds only about
15000 parameters over decoder-only injection (less than 7\%). This shows
that adding the first encoder layer alone approaches full-layer performance at
almost no extra parameter cost, serving as a cost-effective compromise for
feature extraction. Moreover, both configurations add only 8.1\% and 5.4\% of
the backbone parameters (4434595), respectively, keeping the overhead
minimal. Overall, the adopted configuration offers the best balance between
parameter overhead and performance. The results are reported in Tables~\ref{tab:ixi-lora-position-ablation} and~\ref{tab:fastmri-lora-position-ablation}.

\subsubsection{Ablation of Dynamic Adaptation Components}

\begin{table}[t]
  \caption{Module ablation study on the IXI dataset under different acceleration factors}
  \label{tab:ixi-module-ablation}
  \centering
  \scriptsize
  \renewcommand{\arraystretch}{1.12}
  \begin{tabular*}{\linewidth}{@{\extracolsep{\fill}}llcc@{}}
    \toprule
    Module configuration & Factor & PSNR & SSIM \\
    \midrule
    Coarse partitioning & $4\times$ & $42.17 \pm 5.662$ & $0.9881 \pm 0.00856$ \\
    & $8\times$ & $34.01 \pm 5.420$ & $0.9532 \pm 0.02138$ \\
    & $16\times$ & $29.94 \pm 5.406$ & $0.8970 \pm 0.03952$ \\
    \midrule
    \shortstack[l]{AB fine-grained partitioning}
      & $4\times$ & $42.16 \pm 5.668$ & $0.9881 \pm 0.00856$ \\
    & $8\times$ & $34.03 \pm 5.381$ & $0.9534 \pm 0.02114$ \\
    & $16\times$ & $29.97 \pm 5.376$ & $0.8976 \pm 0.03893$ \\
    \bottomrule
  \end{tabular*}
\end{table}

To verify the effectiveness of each component of Shared LoRA, we incrementally
remove or replace the relevant modules while keeping the LoRA insertion
locations and all other experimental conditions fixed. Compared with coarse
single-gate partitioning, AB fine-grained gating yields PSNR gains of no more
than 0.03 dB across all factors (34.03 vs. 34.01 dB at \(8\times\) and 29.97
vs. 29.94 dB at \(16\times\), with \(4\times\) nearly identical), far below the
reported standard deviations and at the cost of additional computation,
indicating that fine-grained per-branch gating offers negligible benefit and
coarse gating suffices. The results are reported in Table~\ref{tab:ixi-module-ablation}.

\section{Conclusion}

This paper proposes Shared LoRA, which freezes the pretrained SHFormer
backbone, incorporates shared low-rank adaptation branches into convolutional
layers, and uses an acceleration-factor-conditioned GateNet to dynamically
modulate the LoRA residual strength of each layer, thereby reusing a single set
of parameters across multiple acceleration factors. During training,
acceleration factors and sampling masks are randomly sampled. During inference,
the adapter response is dynamically adjusted according to the target
acceleration factor. Experiments show that the method outperforms other
methods across multiple acceleration factors on IXI-T2 and fastMRI, alleviates
task interference in joint multi-factor training, and maintains stable
performance on unseen but adjacent acceleration factors. Ablation studies
further indicate that coarse gating is sufficient for effective adaptation,
providing a practical balance between reconstruction quality, parameter
efficiency, and computational cost.

\begin{acks}
This work was supported by the National Natural Science
Foundation of China (No.62306320, 61976217), the Open
Project Program of State Key Lab. for Novel Software Tech-
nology (No. KFKT2024B32), and the Natural Science Foun-
dation of Jiangsu Province (No. BK20231063)
\end{acks}

\bibliographystyle{ACM-Reference-Format}
\bibliography{sample-base}

@String{Computing = "Computing" }

@String{Computer = "{IEEE} Computer" }

@String{Springer = "Springer-Verlag" }

@ArtifactSoftware{R,
    title = {R: A Language and Environment for Statistical Computing},
    author = {{R Core Team}},
    organization = {R Foundation for Statistical Computing},
    address = {Vienna, Austria},
    year = {2019},
    url = {https://www.R-project.org/},
}

@article{cukur2026mri,
  author = {Cukur, Tolga and Dar, Salman U. H. and Nezhad, Valiyeh A. and Jun, Yohan and Kim, Tae Hyung and Fujita, Shohei and Bilgic, Berkin},
  title = {A Tutorial on MRI Reconstruction: From Modern Methods to Clinical Implications},
  journal = {IEEE Transactions on Biomedical Engineering},
  volume = {73},
  number = {5},
  pages = {1900--1920},
  year = {2026},
  doi = {10.1109/TBME.2025.3617575}
}

@inproceedings{yaman2020self,
  author = {Yaman, Burhaneddin and Hosseini, Seyed Amir Hossein and Moeller, Steen and Ellermann, Jutta and Ugurbil, Kamil and Akcakaya, Mehmet},
  title = {Self-Supervised Physics-Based Deep Learning MRI Reconstruction Without Fully-Sampled Data},
  booktitle = {2020 IEEE 17th International Symposium on Biomedical Imaging (ISBI)},
  pages = {921--925},
  year = {2020},
  organization = {IEEE},
  doi = {10.1109/isbi45749.2020.9098514},
  url = {https://doi.org/10.1109/ISBI45749.2020.9098514}
}

@article{heckel2024deep,
  author = {Heckel, Reinhard and Jacob, Mathews and Chaudhari, Akshay and Perlman, Or and Shimron, Efrat},
  title = {Deep Learning for Accelerated and Robust MRI Reconstruction},
  journal = {Magnetic Resonance Materials in Physics, Biology and Medicine},
  volume = {37},
  number = {3},
  pages = {335--368},
  year = {2024},
  doi = {10.1007/s10334-024-01173-8}
}

@article{lustig2007sparse,
  author = {Lustig, Michael and Donoho, David and Pauly, John M.},
  title = {Sparse MRI: The Application of Compressed Sensing for Rapid MR Imaging},
  journal = {Magnetic Resonance in Medicine},
  volume = {58},
  number = {6},
  pages = {1182--1195},
  year = {2007},
  doi = {10.1002/mrm.21391}
}

@article{schlemper2018deep,
  author = {Schlemper, Jo and Caballero, Jose and Hajnal, Joseph V. and Price, Anthony N. and Rueckert, Daniel},
  title = {A Deep Cascade of Convolutional Neural Networks for Dynamic MR Image Reconstruction},
  journal = {IEEE Transactions on Medical Imaging},
  volume = {37},
  number = {2},
  pages = {491--503},
  year = {2018},
  doi = {10.1109/TMI.2017.2760978}
}

@article{aggarwal2019modl,
  author = {Aggarwal, Hemant K. and Mani, Merry P. and Jacob, Mathews},
  title = {MoDL: Model-Based Deep Learning Architecture for Inverse Problems},
  journal = {IEEE Transactions on Medical Imaging},
  volume = {38},
  number = {2},
  pages = {394--405},
  year = {2019},
  doi = {10.1109/tmi.2018.2865356},
  url = {https://doi.org/10.1109/tmi.2018.2865356}
}

@inbook{sriram2020varnet,
  author = {Sriram, Anuroop and Zbontar, Jure and Murrell, Tullie and Defazio, Aaron and Zitnick, C. Lawrence and Yakubova, Nafissa and Knoll, Florian and Johnson, Patricia},
  title = {End-to-End Variational Networks for Accelerated MRI Reconstruction},
  booktitle = {Medical Image Computing and Computer Assisted Intervention -- MICCAI 2020},
  pages = {64--73},
  year = {2020},
  publisher = {Springer International Publishing},
  doi = {10.1007/978-3-030-59713-9_7},
  url = {https://doi.org/10.1007/978-3-030-59713-9_7}
}

@article{hammernik2018variational,
  author = {Hammernik, Kerstin and Klatzer, Teresa and Kobler, Erich and Recht, Michael P. and Sodickson, Daniel K. and Pock, Thomas and Knoll, Florian},
  title = {Learning a Variational Network for Reconstruction of Accelerated MRI Data},
  journal = {Magnetic Resonance in Medicine},
  volume = {79},
  number = {6},
  pages = {3055--3071},
  year = {2018},
  doi = {10.1002/mrm.26977}
}

@article{noordman2023complexities,
  author = {Noordman, Constant Richard and Yakar, Derya and Bosma, Joeran and Simonis, Frank Frederikus Jacobus and Huisman, Henkjan},
  title = {Complexities of Deep Learning-Based Undersampled MR Image Reconstruction},
  journal = {European Radiology Experimental},
  volume = {7},
  number = {1},
  pages = {58},
  year = {2023},
  doi = {10.1186/s41747-023-00372-7}
}

@article{qiu2024multicontrast,
  author = {Qiu, Yiran and Zhang, Haotian and Ma, Qiaoyu and Yang, Guangsong and Lai, Zongying},
  title = {Multi-Contrast MRI Reconstruction Based on Frequency Domain Separation and Cross-Self-Attention},
  journal = {IEEE Access},
  volume = {12},
  pages = {55062--55076},
  year = {2024},
  doi = {10.1109/ACCESS.2024.3388379}
}

@article{guo2024reconformer,
  author = {Guo, Pengfei and Mei, Yiqun and Zhou, Jinyuan and Jiang, Shanshan and Patel, Vishal M.},
  title = {ReconFormer: Accelerated MRI Reconstruction Using Recurrent Transformer},
  journal = {IEEE Transactions on Medical Imaging},
  volume = {43},
  number = {1},
  pages = {582--593},
  year = {2024},
  doi = {10.1109/tmi.2023.3314747},
  url = {https://doi.org/10.1109/tmi.2023.3314747}
}

@article{ramanarayanan2025shformer,
  author = {Ramanarayanan, Sriprabha and Rahul, G. S. and Al Fahim, Mohammad and Ram, Keerthi and Venkatesan, Ramesh and Sivaprakasam, Mohanasankar},
  title = {SHFormer: Dynamic Spectral Filtering Convolutional Neural Network and High-Pass Kernel Generation Transformer for Adaptive MRI Reconstruction},
  journal = {Neural Networks},
  volume = {187},
  pages = {107334},
  year = {2025},
  doi = {10.1016/j.neunet.2025.107334}
}

@inproceedings{houlsby2019adapter,
  author = {Houlsby, Neil and Giurgiu, Andrei and Jastrzebski, Stanislaw and Morrone, Bruna and De Laroussilhe, Quentin and Gesmundo, Andrea and Attariyan, Mona and Gelly, Sylvain},
  title = {Parameter-Efficient Transfer Learning for NLP},
  booktitle = {Proceedings of the 36th International Conference on Machine Learning},
  series = {Proceedings of Machine Learning Research},
  volume = {97},
  pages = {2790--2799},
  year = {2019},
  publisher = {PMLR},
  url = {https://proceedings.mlr.press/v97/houlsby19a.html}
}

@inproceedings{hu2022lora,
  author = {Hu, Edward J. and Shen, Yelong and Wallis, Phillip and Allen-Zhu, Zeyuan and Li, Yuanzhi and Wang, Shean and Wang, Lu and Chen, Weizhu},
  title = {LoRA: Low-Rank Adaptation of Large Language Models},
  booktitle = {International Conference on Learning Representations},
  year = {2022},
  url = {https://iclr.cc/virtual/2022/poster/6319}
}

@inproceedings{yeh2024navigating,
  author = {Yeh, Shih-Ying and Hsieh, Yu-Guan and Gao, Zhidong and Yang, Bernard B. W. and Oh, Giyeong and Gong, Yanmin},
  title = {Navigating Text-To-Image Customization: From LyCORIS Fine-Tuning to Model Evaluation},
  booktitle = {International Conference on Learning Representations},
  year = {2024},
  url = {https://iclr.cc/virtual/2024/poster/17484}
}

@inproceedings{chen2022adaptformer,
  author = {Chen, Shoufa and Ge, Chongjian and Tong, Zhan and Wang, Jiangliu and Song, Yibing and Wang, Jue and Luo, Ping},
  title = {AdaptFormer: Adapting Vision Transformers for Scalable Visual Recognition},
  booktitle = {Advances in Neural Information Processing Systems 35},
  pages = {16664--16678},
  year = {2022},
  series = {NeurIPS 2022},
  publisher = {Neural Information Processing Systems Foundation, Inc. (NeurIPS)},
  doi = {10.52202/068431-1212},
  url = {https://doi.org/10.52202/068431-1212}
}

@article{wang2025pnp,
  author = {Wang, Jianmin and Liu, Chunyan and Zhong, Yuxiang and Liu, Xinling and Wang, Jianjun},
  title = {Deep plug-and-play MRI reconstruction based on multiple complementary priors},
  journal = {Magnetic Resonance Imaging},
  volume = {115},
  pages = {110244},
  year = {2025},
  doi = {10.1016/j.mri.2024.110244},
  url = {https://www.sciencedirect.com/science/article/pii/S0730725X2400225X}
}

@article{huang2024dured,
  author = {Huang, Peizhou and Zhang, Chaoyi and Zhang, Xiaoliang and Li, Xiaojuan and Dong, Liang and Ying, Leslie},
  title = {Self-Supervised Deep Unrolled Reconstruction Using Regularization by Denoising},
  journal = {IEEE Transactions on Medical Imaging},
  volume = {43},
  number = {3},
  pages = {1203--1213},
  year = {2024},
  doi = {10.1109/TMI.2023.3332614}
}

@article{yi2023fmt,
  author = {Yi, Qiaosi and Fang, Faming and Zhang, Guixu and Zeng, Tieyong},
  title = {Frequency Learning via Multi-Scale Fourier Transformer for MRI Reconstruction},
  journal = {IEEE Journal of Biomedical and Health Informatics},
  volume = {27},
  number = {11},
  pages = {5506--5517},
  year = {2023},
  doi = {10.1109/JBHI.2023.3311189}
}

@article{xu2026hieradaptmr,
  author = {Xu, Ruru and Oksuz, Ilkay},
  title = {HierAdaptMR: Cross-Center Cardiac MRI Reconstruction with Hierarchical Feature Adapters},
  journal = {Journal of Cardiovascular Magnetic Resonance},
  volume = {28},
  pages = {102490},
  year = {2026},
  note = {Proceedings of the SCMR 29th Annual Scientific Sessions, Rio de Janeiro, Brazil, 4--7 February, 2026},
  doi = {10.1016/j.jocmr.2025.102490},
  url = {https://www.sciencedirect.com/science/article/pii/S1097664725006520}
}

@article{geng2026ttpssfl,
  author = {Geng, Chenghu and Jiang, Mingfeng and Ruan, Dongsheng and Yu, Chengjin and Sun, Hong and Liu, Feng and Xia, Ling and Yang, Guang},
  title = {TTP-SSFL: Test-Time Personalization Self-Supervised Federated Learning for Accelerating MR Image Reconstruction},
  journal = {IEEE Transactions on Neural Networks and Learning Systems},
  pages = {1--12},
  year = {2026},
  doi = {10.1109/TNNLS.2026.3703424}
}

@inbook{lyu2024upcmr,
  author = {Lyu, Donghang and Rao, Chinmay and Staring, Marius and van Osch, Matthias J. P. and Doneva, Mariya and Lamb, Hildo J. and Pezzotti, Nicola},
  title = {UPCMR: A Universal Prompt-Guided Model for Random Sampling Cardiac MRI Reconstruction},
  booktitle = {Statistical Atlases and Computational Models of the Heart. Workshop, CMRxRecon and MBAS Challenge Papers},
  pages = {453--463},
  year = {2025},
  publisher = {Springer Nature Switzerland},
  doi = {10.1007/978-3-031-87756-8_44},
  url = {https://doi.org/10.1007/978-3-031-87756-8_44}
}

@inproceedings{ren2024melora,
  author = {Ren, Pengjie and Shi, Chengshun and Wu, Shiguang and Zhang, Mengqi and Ren, Zhaochun and de Rijke, Maarten and Chen, Zhumin and Pei, Jiahuan},
  title = {MELoRA: Mini-Ensemble Low-Rank Adapters for Parameter-Efficient Fine-Tuning},
  booktitle = {Proceedings of the 62nd Annual Meeting of the Association for Computational Linguistics (Volume 1: Long Papers)},
  pages = {3052--3064},
  year = {2024},
  publisher = {Association for Computational Linguistics},
  doi = {10.18653/v1/2024.acl-long.168},
  url = {https://doi.org/10.18653/v1/2024.acl-long.168}
}

@inproceedings{pmlr-v235-liu24bn,
  author = {Liu, Shih-Yang and Wang, Chien-Yi and Yin, Hongxu and Molchanov, Pavlo and Wang, Yu-Chiang Frank and Cheng, Kwang-Ting and Chen, Min-Hung},
  title = {{DoRA}: Weight-Decomposed Low-Rank Adaptation},
  booktitle = {Proceedings of the 41st International Conference on Machine Learning},
  pages = {32100--32121},
  year = {2024},
  editor = {Salakhutdinov, Ruslan and Kolter, Zico and Heller, Katherine and Weller, Adrian and Oliver, Nuria and Scarlett, Jonathan and Berkenkamp, Felix},
  volume = {235},
  series = {Proceedings of Machine Learning Research},
  publisher = {PMLR},
  url = {https://proceedings.mlr.press/v235/liu24bn.html}
}

@article{eo2018kikinet,
  author = {Eo, Taejoon and Jun, Yohan and Kim, Taeseong and Jang, Jinseong and Lee, Ho-Joon and Hwang, Dosik},
  title = {KIKI-net: Cross-Domain Convolutional Neural Networks for Reconstructing Undersampled Magnetic Resonance Images},
  journal = {Magnetic Resonance in Medicine},
  volume = {80},
  number = {5},
  pages = {2188--2201},
  year = {2018},
  doi = {10.1002/mrm.27201},
  url = {https://doi.org/10.1002/mrm.27201}
}

@article{yang2018dagan,
  author = {Yang, Guang and Yu, Simiao and Dong, Hao and Slabaugh, Greg and Dragotti, Pier Luigi and Ye, Xujiong and Liu, Fangde and Arridge, Simon and Keegan, Jennifer and Guo, Yike and Firmin, David},
  title = {DAGAN: Deep De-Aliasing Generative Adversarial Networks for Fast Compressed Sensing MRI Reconstruction},
  journal = {IEEE Transactions on Medical Imaging},
  volume = {37},
  number = {6},
  pages = {1310--1321},
  year = {2018},
  doi = {10.1109/TMI.2017.2785879},
  url = {https://doi.org/10.1109/TMI.2017.2785879}
}

\end{document}